\documentclass{article}
\usepackage{spconf,amsmath,graphicx}
\usepackage{booktabs, multirow, xcolor}
\usepackage{cuted}  
\usepackage{caption}
\usepackage{xcolor}
\usepackage{amssymb}
\usepackage{subcaption} 
\usepackage{verbatim}
\usepackage[pagebackref,breaklinks,colorlinks,citecolor=blue]{hyperref}

\title{Quantifying the sampling frequency invariance of time-series foundation models for sensor data applications}
\name{\begin{tabular}[t]{@{}c@{}}
Sourish Gunesh Dhekane$^{2}$, Abinaya Kumar$^{1}$, Torsten Woertwein$^{1}$, \\
Yan-Bo Lin$^{1}$, Thomas Ploetz$^{2}$, Shabnam Ghaffarzadegan$^{1}$
\end{tabular}}
\address{$^{1}$Bosch Research, USA - Bosch Center for Artificial Intelligence \\
         $^{2}$School of Interactive Computing, Georgia Institute of Technology, USA}
\begin{document}
\ninept
\maketitle
\newcommand{\torsten}[1]{{{\textcolor{purple}{[#1 -- TW]}}}}
\newcommand{\yb}[1]{{{\textcolor{blue}{[#1 -- YB]}}}}
\newcommand{\ak}[1]{{{\textcolor{red}{[#1 -- AK]}}}}
\setlength{\stripsep}{2pt}   

\vspace{-6pt}
\begin{abstract}
\vspace{-1pt}
Time-Series Foundation Models (TSFMs) are increasingly used in sensor applications, where sampling frequency (SF) varies across sensors, devices, and operating modes (e.g., power saving).
Ideally, TSFMs must remain effective during such SF changes, provided the SF retains task-critical bandwidth.
In this paper, we stress-test 4 prominent TSFMs: MOMENT, Chronos-2, Moirai-MoE, and TimesFM, to quantify their SF-invariance and investigate whether their behavior tracks the resulting loss of information bandwidth rather than the SF value itself.
Our stress-test probes 3 dimensions: (\textit{i}) the shift in embeddings, (\textit{ii}) the degradation in forecasting performance, and (\textit{iii}) the degradation in downstream performance, quantified by three novel metrics: FSI, FDI, and CDI, respectively.  
To our knowledge, this is the first study to place SF-invariance at the core of TSFM evaluations.
Our results characterize: (\textit{i}) how SF-invariance depends on the margin between signal's native-SF and the bandwidth its task requires, (\textit{ii}) how much of that invariance the choice of TSFM explains, and (\textit{iii}) how much of TSFM's performance degradation under a lower SF reflects information loss rather than a limitation of their abilities.
These findings reframe SF-invariance as a measurable, signal-aware property of the (signal$\times$TSFM) pair, and guide practitioners on when TSFMs display SF-invariance.
\end{abstract}
\vspace{-1pt}
\vspace{-6pt}
\section{Introduction}
\label{sec:intro}
\vspace{-4pt}
TSFMs are increasingly applied to sensor applications ranging from activity monitoring and healthcare to industrial settings \cite{luo2026toward, kottapalli2025foundation}, where heterogeneous sensors and devices operate at different SFs.
Even a single device changes its SF across operating modes to conserve battery or to optimize bandwidth usage. 
An ideal TSFM should therefore remain effective across all such settings with varying SFs.
A primary obstacle to this is the SF mismatch between TSFM's pretraining and the downstream sensor data, as most pretraining benchmarks span SFs from seconds to years \cite{khanal2026bridging}.
Consequently, at deployment, TSFMs can encounter SFs that differ substantially from those seen in pretraining, yet we have little evidence of how robustly a TSFM behaves under such unseen SF changes.

\begin{figure}[t]
  \centering
  \includegraphics[width=\linewidth]{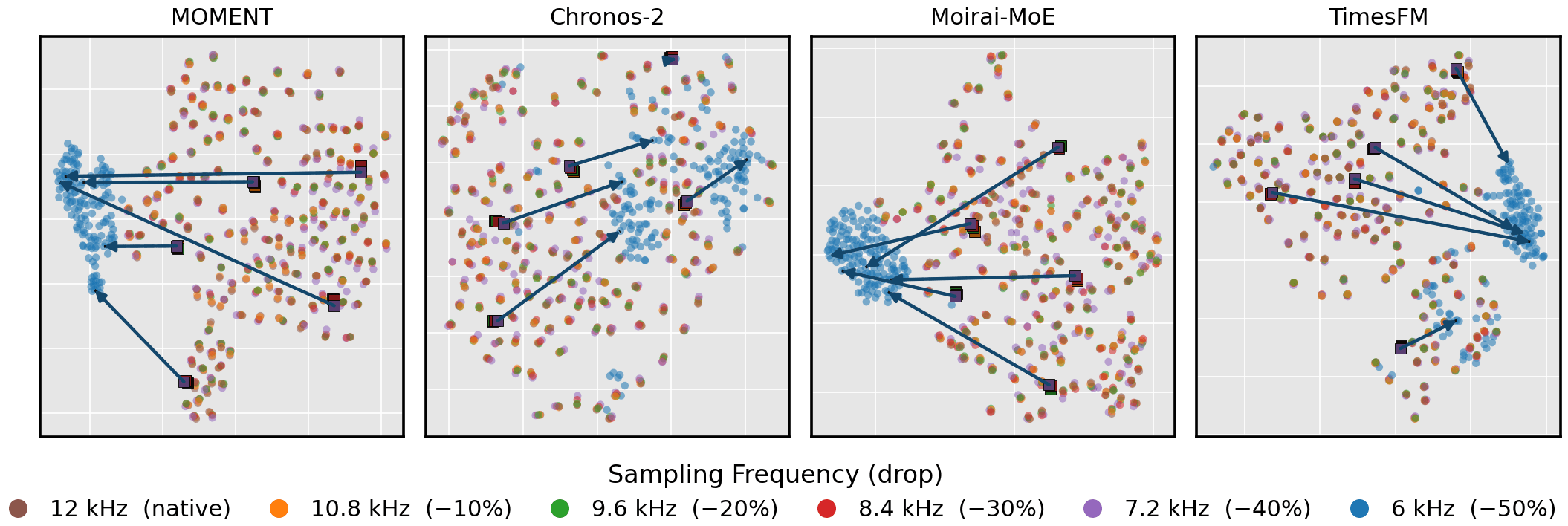}
  \caption{ 
  TSFM embeddings on CWRU vibration data remain stable up to a 40\% reduction in SF, but shift sharply between 40\% and 50\%. In this paper, we study and quantify these behavioral shifts across TSFM embeddings, forecasts, and downstream performance.
}
  \label{fig:motivation_1}
  \vspace{-14pt}
\end{figure}

However, a behavioral change in TSFMs may sometimes be inevitable. 
By the Nyquist--Shannon sampling theorem \cite{shannon1949communication}, reducing the SF removes signal content above the new Nyquist frequency, so a lower SF entails some information loss, and a model's behavior is therefore expected to change.
What should distinguish a SF-invariant TSFM is that this change stays proportional to the information loss, and that for a downstream task, the performance should be robust as long as the task-critical information is still retained in the signal.
Also, disproportionate changes (See Figure \ref{fig:motivation_1}) can make the underlying application less robust.
While some existing works in the literature have pointed out these issues \cite{wang2025frequency,gupta2025evaluating}, a systematic stress test that isolates the aspect of SF in TSFMs and quantifies their robustness to varying SF
is absent from the literature.

In this paper, we perform such a systematic stress test on 4 prominent TSFMs used in the community: MOMENT \cite{goswami2024moment}, Chronos-2 \cite{ansari2025chronos}, Moirai-MoE \cite{liu2024moirai}, and TimesFM \cite{das2024decoder}, on use cases that span accelerometers, gyroscope, vibration, and ECG sensors.
Our stress test covers 3 dimensions of TSFMs that track a change in: embeddings, forecasting performance, and downstream performance, upon a change in SF. 
To quantify this change, we propose the metrics of Frequency Sensitivity Index (FSI), Forecast Degradation Index (FDI), and Classification Degradation Index (CDI), respectively. 
The results of our stress test reveal interesting findings: 
(\textit{i}) the 3 dimensions do not respond to SF-reduction at the same point: while the embeddings begin drifting immediately, forecasting and downstream performances do not degrade much until the SF approaches the task-critical bandwidth,
(\textit{ii}) SF-invariance is influenced more by the gap between native-SF and the information bandwidth critical for the downstream tasks as compared to the choice of the TSFM itself,
and (\textit{iii}) however, all TSFMs have scope to improve their SF-invariance, as their downstream performance drops further than the information loss accounted for by SF-degradation.
Together, these findings give practitioners a basis for choosing design choices (e.g., a sensor's SF, choice of TSFM, etc.) in their sensor applications and motivate the community to treat SF-invariance as an integral part of TSFM evaluation, eventually towards improving it.
\vspace{-6pt}
\section{Stress Test Design}
\label{sec:stress}
\vspace{-3pt}
\begin{table}[t]
\centering\setlength{\tabcolsep}{3pt}
\resizebox{\columnwidth}{!}{%
\begin{tabular}{llll}
\toprule
TSFM & Backbone & Majority Pre-train SF & Pre-train loss \\
\midrule
MOMENT~\cite{goswami2024moment}    & enc.-only (T5)  & $\ll\!1$\,Hz & masked recon.\ (MSE) \\
Chronos-2~\cite{ansari2025chronos} & enc.-only\ (T5) & $\ll\!1$\,Hz & quantile (pinball) \\
Moirai-MoE~\cite{liu2024moirai}    & dec.-only, MoE  & $\ll\!1$\,Hz & distrib.\ NLL \\
TimesFM~\cite{das2024decoder}      & dec.-only       & $\ll\!1$\,Hz & next-patch (MSE) \\
\bottomrule
\end{tabular}%
}
\caption{We use MOMENT-1-large, Chronos-2, Moirai-MoE-1.0-R-small, and TimesFM-1.0-200M TSFMs having most pre-training data $\ll\!1$ Hz; Chronos-2 \& TimesFM also use SF-free synthetic data. 
}
\vspace{-14pt}
\label{tab:tsfm}
\end{table}
We stress-test 4 prominent TSFMs spanning the transformer-based architecture family (Table \ref{tab:tsfm}): encoder-only MOMENT~\cite{goswami2024moment} and Chronos-2~\cite{ansari2025chronos}, decoder-only TimesFM~\cite{das2024decoder} and Mixture-of-Experts Moirai-MoE~\cite{liu2024moirai}. 
Although non-transformer TSFMs \cite{auer2026tirex, bhethanabhotla2024mamba4cast} are also actively explored (even designed to be SF-equivariant \cite{graf2025flowstate}), they do not match the widespread adoption of transformer-based architectures due to their parallelization-friendliness, scalability, and efficiency in learning from large datasets \cite{liang2024foundation}.
Hence, we focus on transformer-based TSFMs. 

Our stress-test probes 3 dimensions. 
Dimension~1 measures the shift in the internal embedding space, on which the
downstream use of TSFM depends.
However, it alone cannot say whether, and to what extent, that shift would matter for a downstream task.
Hence, we complement it with Dimension~2 that measures the degradation in forecasting (i.e., TSFM's pre-training objective) and Dimension~3 that measures the degradation on a downstream classification task, covering a majority of TSFM deployments on sensor data.
These 3 dimensions ensure a holistic conclusion on SF-invariance.
\vspace{-7pt}
\subsection{Common Setup and Quantification Method}
\label{sec:common}
Let $\mathbf{x}\in\mathbb{R}^{L\times C}$ be a $C$-channel window acquired at the sensor's native SF $F$, with $L$ set to the context size of the TSFM (e.g., HHAR accelerometer windows fed to MOMENT have $C=3$, $L=512$, and $F=100$ Hz). 
Each frozen TSFM defines an embedding map
$E:\mathbb{R}^{L\times C}\!\to\!\mathbb{R}^{d}$, whose output $E(\mathbf{x})$ is
used by Dimensions~1 and~3, and a forecast, used by Dimension~2.
To emulate a sensor with SF $f\le F$, we use band-limiting operator $\mathcal{D}_\rho$, indexed by the fractional SF drop $\rho=1-f/F$.
It removes the spectral content above the target Nyquist frequency $f/2$ while returning a window of the original length $L$.
It does so by performing anti-aliasing decimation of $\mathbf{x}$ to have $n_f=\mathrm{round}(Lf/F)$ samples, which a sensor with SF $f$ would have produced, and then interpolating back to $L$ with a polyphase filter. 
Here, the sequence length and token count are identical for every $\rho$; only the effective bandwidth of the input changes.
We keep L fixed on purpose, as TSFMs are sensitive to the input length of the data \cite{nochumsohn2025super, zukowska2024towards}, irrespective of what SF it comes from, and hence, changing $L$ confounds the effect of SF with the input length, which we avoid with our design choice. 
We write $\mathcal{D}_0\mathbf{x}=\mathbf{x}$ and call $\mathcal{D}_\rho\mathbf{x}$ the band-limited input.
Our stress test therefore measures SF-invariance in terms of effective bandwidth at a fixed token budget. 
This matches deployments where low-SF streams are resampled to the model's input rate before inference. 
Feeding the $n_f$ samples directly would also change the time span covered by each patch, which is a separate effect and we leave it to future work. 

Our stress test uses a common quantification method for each dimension,
where a discrepancy $\delta(\rho)$ between the model's band-limited and native responses is measured, with $\delta(0)=0$.
To visualize this discrepancy, we plot the $\delta$-versus-$\rho$ curve.
This curve is quantified by calculating the area under it on a single grid $\rho_j=j\,\Delta\rho$, $j=0,\dots,J$, with step $\Delta\rho=0.1$ and $J=9$, so the SF is swept from native ($\rho_0=0$) down to $10\%$ of native ($\rho_J=0.9$).
\begin{equation}
  \mathcal{I}=\int_{0}^{\rho_{\max}}\!\!\delta(\rho)\,\mathrm{d}\rho
  \;\approx\;\sum_{j=1}^{J}\tfrac{1}{2}
  \big(\delta(\rho_{j-1})+\delta(\rho_{j})\big)\,(\rho_j-\rho_{j-1})
  \label{eq:index}
\end{equation}
A smaller index $\mathcal{I}$ means less behavioral drift upon SF-reduction, hence greater SF-invariance. 
We compute this $\mathcal{I}$ on a non-overlapping set of length-$L$ windows $\mathcal{W}$, stratified across recording groups (subject or device) and classes with a fixed per-group quota ($\ge\!20$) and a shared seed, so every TSFM is evaluated on identical windows. 
\vspace{-9pt}
\subsection{Dimension 1: Frequency Sensitivity Index (FSI)}
\vspace{-2pt}
\label{sec:dim1}
To quantify Dimension 1, we measure cosine distance between the native and band-limited embeddings, defined as:
\begin{equation}
  \delta^{\mathrm{emb}}(\rho)=
  1-\frac{\big\langle E(\mathbf{x}),\,E(\mathcal{D}_{\rho}\mathbf{x})\big\rangle}
         {\big\|E(\mathbf{x})\big\|\,\big\|E(\mathcal{D}_{\rho}\mathbf{x})\big\|}.
  \label{eq:drift}
\end{equation}
We use the cosine distance as our discrepancy measure, as it is invariant to the
representation scale of TSFM backbones.
FSI is calculated by applying Eq.~\ref{eq:index} to $\delta^{\mathrm{emb}}$, and taking mean$\pm$std over $\mathcal{W}$. 
\vspace{-9pt}
\subsection{Dimension 2: Forecast Degradation Index (FDI)}
\label{sec:dim2}
\vspace{-2pt}
For quantifying Dimension 2, we first define our forecasting task under SF change.
At the native SF, the TSFM is required to forecast $H$ samples given a context of $L$ samples.
At reduced SFs, we keep the same forecast requirement of $H$ samples (at the native SF) but given a band-limited context: the TSFM must predict the same native-SF forecast $\mathbf{y}\in\mathbb{R}^{H\times C}$ from $\mathcal{D}_\rho\mathbf{x}$ rather than from $\mathbf{x}$. 
The target $\mathbf{y}$ is held at the native SF for every $\rho$.
This is because the goal is to address the deployment-relevant question of whether the TSFM reading a band-limited input can still predict the native-SF horizon.

Each model's native output (Moirai-MoE's sample paths,
Chronos-2's quantiles, TimesFM's point-quantile head) is collapsed to a common
point forecast and scored by the normalized RMSE $e(\rho)$, defined as the RMSE against the
native target divided by that target's standard deviation. 
Based on this response, we then define the discrepancy as the increase in this error over the native forecast, $\delta^{\mathrm{fc}}(\rho)=e(\rho)-e(0)$, and FDI is calculated by applying Eq.~\ref{eq:index} to it at $H=128$ native samples ($32$ for the short pre-segmented ECG beats). 
We exclude MOMENT from this dimension as it does not have a pretrained forecasting head.
In our experiments, we observe that the error distribution is heavy-tailed for all three TSFMs; hence, we aggregate FDI with the median and report the IQR.

\begin{table*}[t]
\centering\footnotesize\setlength{\tabcolsep}{4.5pt}
\begin{tabular}{llccccccc}
\toprule
\multicolumn{2}{r}{Datasets\,$\rightarrow$} & \multicolumn{4}{c}{HHAR} & \multicolumn{2}{c}{Bearing} & MIT-BIH\\
\cmidrule(lr){3-6}\cmidrule(lr){7-8}\cmidrule(lr){9-9}
Dim & TSFM\,$\downarrow$ & Acc 100Hz & Acc 200Hz & Gyr 100Hz & Gyr 200Hz & Vib 12kHz & Vib 48kHz & ECG 125Hz\\
\midrule
\multirow{4}{*}{\shortstack{(1)\\FSI$\downarrow$\\{\scriptsize mean$\pm$std}}} & Chronos-2 & 0.061\,$\pm$\,0.059 & 0.068\,$\pm$\,0.045 & 0.026\,$\pm$\,0.016 & 0.033\,$\pm$\,0.021 & 0.092\,$\pm$\,0.023 & 0.027\,$\pm$\,0.017 & 0.125\,$\pm$\,0.064\\
 & Moirai-MoE & 0.023\,$\pm$\,0.025 & 0.036\,$\pm$\,0.031 & 0.011\,$\pm$\,0.012 & 0.015\,$\pm$\,0.017 & 0.098\,$\pm$\,0.015 & 0.006\,$\pm$\,0.002 & 0.022\,$\pm$\,0.011\\
 & MOMENT & 0.023\,$\pm$\,0.026 & 0.028\,$\pm$\,0.033 & 0.008\,$\pm$\,0.009 & 0.004\,$\pm$\,0.007 & 0.116\,$\pm$\,0.026 & 0.006\,$\pm$\,0.001 & 0.012\,$\pm$\,0.006\\
 & TimesFM & 0.032\,$\pm$\,0.066 & 0.031\,$\pm$\,0.036 & 0.013\,$\pm$\,0.014 & 0.016\,$\pm$\,0.015 & 0.051\,$\pm$\,0.042 & 0.006\,$\pm$\,0.006 & 0.027\,$\pm$\,0.027\\
\midrule
\multirow{3}{*}{\shortstack{(2)\\FDI$\downarrow$\\{\scriptsize median (IQR)}}} & Chronos-2 & 0.439\,$\pm$\,0.597 & 0.655\,$\pm$\,0.750 & 0.253\,$\pm$\,0.680 & 0.260\,$\pm$\,1.279 & 0.079\,$\pm$\,0.332 & 0.142\,$\pm$\,0.451 & 4.387\,$\pm$\,10.474\\
 & Moirai-MoE & 1.953\,$\pm$\,3.973 & 1.817\,$\pm$\,4.328 & 0.784\,$\pm$\,2.445 & 0.515\,$\pm$\,3.103 & 0.079\,$\pm$\,0.834 & 0.074\,$\pm$\,0.542 & 5.692\,$\pm$\,12.137\\
 & TimesFM & 0.484\,$\pm$\,0.727 & 0.614\,$\pm$\,1.032 & 0.102\,$\pm$\,0.821 & 0.164\,$\pm$\,1.502 & 0.036\,$\pm$\,0.116 & -0.011\,$\pm$\,0.397 & 1.942\,$\pm$\,8.741\\
\midrule
\multirow{8}{*}{\shortstack{(3)\\CDI$\downarrow$\\{\scriptsize mean$\pm$std}}} & Chronos-2 & 0.073\,$\pm$\,0.057 & 0.085\,$\pm$\,0.045 & 0.110\,$\pm$\,0.048 & 0.094\,$\pm$\,0.047 & 0.386\,$\pm$\,0.034 & 0.061\,$\pm$\,0.009 & 0.203\,$\pm$\,0.000\\
 & Moirai-MoE & 0.093\,$\pm$\,0.063 & 0.125\,$\pm$\,0.025 & 0.086\,$\pm$\,0.041 & 0.079\,$\pm$\,0.033 & 0.309\,$\pm$\,0.026 & 0.048\,$\pm$\,0.003 & 0.148\,$\pm$\,0.000\\
 & MOMENT & 0.055\,$\pm$\,0.035 & 0.114\,$\pm$\,0.065 & 0.042\,$\pm$\,0.018 & 0.024\,$\pm$\,0.011 & 0.413\,$\pm$\,0.038 & 0.040\,$\pm$\,0.002 & 0.095\,$\pm$\,0.000\\
 & TimesFM & 0.069\,$\pm$\,0.034 & 0.085\,$\pm$\,0.045 & 0.059\,$\pm$\,0.022 & 0.057\,$\pm$\,0.028 & 0.349\,$\pm$\,0.055 & 0.037\,$\pm$\,0.002 & 0.141\,$\pm$\,0.000\\
\cmidrule(l){2-9}
 & enh-Chronos-2 & \shortstack{0.039\,$\pm$\,0.028 \\ {\scriptsize\textcolor{green!55!black}{($\downarrow$46\%)}}} & \shortstack{0.048\,$\pm$\,0.053 \\ {\scriptsize\textcolor{green!55!black}{($\downarrow$44\%)}}} & \shortstack{0.080\,$\pm$\,0.030 \\ {\scriptsize\textcolor{green!55!black}{($\downarrow$27\%)}}} & \shortstack{0.022\,$\pm$\,0.021 \\ {\scriptsize\textcolor{green!55!black}{($\downarrow$77\%)}}} & \shortstack{0.252\,$\pm$\,0.035 \\ {\scriptsize\textcolor{green!55!black}{($\downarrow$35\%)}}} & \shortstack{0.031\,$\pm$\,0.004 \\ {\scriptsize\textcolor{green!55!black}{($\downarrow$49\%)}}} & \shortstack{0.159\,$\pm$\,0.000 \\ {\scriptsize\textcolor{green!55!black}{($\downarrow$22\%)}}}\\
 & enh-Moirai-MoE & \shortstack{0.110\,$\pm$\,0.060 \\ {\scriptsize\textcolor{red!80!black}{($\uparrow$19\%)}}} & \shortstack{0.085\,$\pm$\,0.017 \\ {\scriptsize\textcolor{green!55!black}{($\downarrow$32\%)}}} & \shortstack{0.052\,$\pm$\,0.015 \\ {\scriptsize\textcolor{green!55!black}{($\downarrow$39\%)}}} & \shortstack{0.030\,$\pm$\,0.020 \\ {\scriptsize\textcolor{green!55!black}{($\downarrow$62\%)}}} & \shortstack{0.267\,$\pm$\,0.049 \\ {\scriptsize\textcolor{green!55!black}{($\downarrow$14\%)}}} & \shortstack{0.034\,$\pm$\,0.003 \\ {\scriptsize\textcolor{green!55!black}{($\downarrow$29\%)}}} & \shortstack{0.107\,$\pm$\,0.000 \\ {\scriptsize\textcolor{green!55!black}{($\downarrow$28\%)}}}\\
 & enh-MOMENT & \shortstack{0.044\,$\pm$\,0.045 \\ {\scriptsize\textcolor{green!55!black}{($\downarrow$20\%)}}} & \shortstack{0.041\,$\pm$\,0.024 \\ {\scriptsize\textcolor{green!55!black}{($\downarrow$64\%)}}} & \shortstack{0.015\,$\pm$\,0.022 \\ {\scriptsize\textcolor{green!55!black}{($\downarrow$65\%)}}} & \shortstack{0.006\,$\pm$\,0.006 \\ {\scriptsize\textcolor{green!55!black}{($\downarrow$77\%)}}} & \shortstack{0.227\,$\pm$\,0.049 \\ {\scriptsize\textcolor{green!55!black}{($\downarrow$45\%)}}} & \shortstack{0.032\,$\pm$\,0.009 \\ {\scriptsize\textcolor{green!55!black}{($\downarrow$20\%)}}} & \shortstack{0.087\,$\pm$\,0.000 \\ {\scriptsize\textcolor{green!55!black}{($\downarrow$8\%)}}}\\
 & enh-TimesFM & \shortstack{0.058\,$\pm$\,0.031 \\ {\scriptsize\textcolor{green!55!black}{($\downarrow$16\%)}}} & \shortstack{0.033\,$\pm$\,0.037 \\ {\scriptsize\textcolor{green!55!black}{($\downarrow$61\%)}}} & \shortstack{0.031\,$\pm$\,0.019 \\ {\scriptsize\textcolor{green!55!black}{($\downarrow$49\%)}}} & \shortstack{0.008\,$\pm$\,0.004 \\ {\scriptsize\textcolor{green!55!black}{($\downarrow$86\%)}}} & \shortstack{0.108\,$\pm$\,0.039 \\ {\scriptsize\textcolor{green!55!black}{($\downarrow$69\%)}}} & \shortstack{0.019\,$\pm$\,0.004 \\ {\scriptsize\textcolor{green!55!black}{($\downarrow$48\%)}}} & \shortstack{0.149\,$\pm$\,0.000 \\ {\scriptsize\textcolor{red!80!black}{($\uparrow$6\%)}}}\\
\bottomrule
\end{tabular}
\caption{FSI, FDI, and CDI of 4 TSFMs on 7 dataset configurations using 5-fold cross-validation; lower value$\implies$more SF-invariance. 
``enh-'' rows are contrastive variants.
Green/red \% values are CDI improvement/degradation.
MIT-BIH uses the specified folds in Dim 3. 
}
\label{tab:stress}
\vspace{-12pt}
\end{table*}
\vspace{-9pt}
\subsection{Dimension 3: Classification Degradation Index (CDI)}
\label{sec:dim3}
To quantify Dimension 3, we measure downstream macro-$F_1$ for the given classification task to be performed across SFs. 
Since TSFMs cannot classify zero-shot, they are probed using a lightweight MLP, which is fitted at the native-SF ($\rho=0$), and evaluated on band-limited inputs under a $\underline{\mathrm{tr}}$ansfer learning scenario, giving $F_1^{\mathrm{tr}}(\rho)$.
For reference, we define an \emph{oracle}
condition, where the probe is refitted at each $\rho$, giving $F_1^{\mathrm{orc}}(\rho)$. 
We define discrepancy as
$\delta^{\mathrm{cls}}(\rho)=F_1^{\mathrm{tr}}(0)-F_1^{\mathrm{tr}}(\rho)$, and CDI is calculated by applying Eq.~\ref{eq:index} to it, as mean$\pm$std over folds $\mathcal{F}$ (random seeds and cross-validation). 
This dimension is the only one that admits the following decomposition, where $F_1^{\mathrm{orc}}(\rho)$ is an upper bound on task-relevant information removed by $\mathcal{D}_\rho$ and the transfer gap $\tau(\rho)$ disentangles information loss and SF-invariance of TSFMs.:
\begin{equation}
  \delta^{\mathrm{cls}}(\rho)=
  \underbrace{\big(F_1^{\mathrm{tr}}(0)-F_1^{\mathrm{orc}}(\rho)\big)}_{\text{information loss}}
  +\underbrace{\big(F_1^{\mathrm{orc}}(\rho)-F_1^{\mathrm{tr}}(\rho)\big)}_{\tau(\rho):\ \text{transfer gap}} .
  \label{eq:cdi-split}
\end{equation}
CDI measures only the SF-invariance and not absolute performance.  
\vspace{-17pt}
\subsection{Contrastive SF-Alignment Head}
\label{sec:adapter}
\vspace{-2pt}
To test whether the transfer gap can be reduced without retraining the TSFM, we freeze its backbone and add a projection head $P$, forming $E_{\mathrm{enh}}(\mathbf{x})=E(\mathbf{x})+P\big(E(\mathbf{x})\big)$.
$P$ is a two-layer MLP trained with an NT-Xent objective whose positives are two band-limited views $(\mathcal{D}_{\rho_1}\mathbf{x},\mathcal{D}_{\rho_2}\mathbf{x})$ of the same window and negatives are the other windows in the batch. 
Here, we sample the two view frequencies $\rho_1,\rho_2$ uniformly from a 10-point grid spanning $\rho \in [0, 0.9]$, and train $P$ for $200$ epochs (NT-Xent temperature $\tau=0.2$, batch size $B=256$, learning rate $10^{-3}$).
$P$ is fit on the training split only, and the Dimension-3 probe is refit on $E_{\mathrm{enh}}$.

\vspace{-7pt}
\section{Results}
\label{sec:results}
\vspace{-4pt}

\begin{figure*}[t]
  \centering
  \begin{subfigure}[b]{0.57\textwidth}
    \centering
    \includegraphics[width=\linewidth]{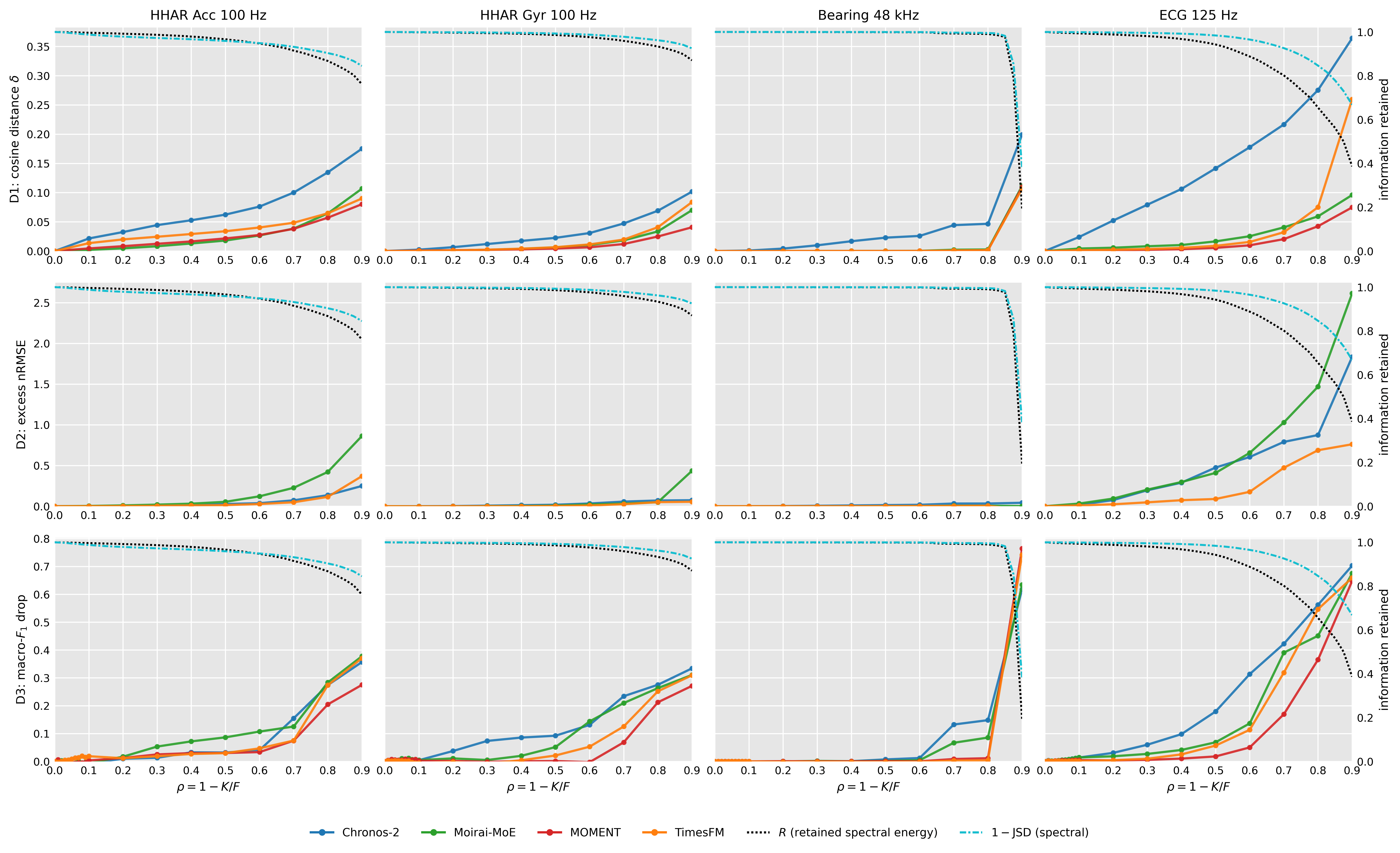}
    \caption{Per-dimension behavior of TSFMs: cosine distance (row 1), excess nRMSE (row 2), and macro-$F_1$ drop (row 3) 
    compared to retained signal information ($R$, $1-$JSD). 
    }
    \label{fig:info-stacked}
  \end{subfigure}
  \hfill
  \begin{subfigure}[b]{0.41\textwidth}
    \centering
    \includegraphics[width=\linewidth]{figures/dim3_levels_mlp_subset_paper.png}
    \caption{$F_1^{\mathrm{tr}}(\rho)$ (red) and $F_1^\mathrm{orc}(\rho)$ (blue) with retained information ($R$, $1-$JSD) upon SF drop. Shaded area is the transfer gap.}
    \label{fig:levels-subset}
  \end{subfigure}
  \caption{Disentangling information loss from SF-invariance. Part (\subref{fig:info-stacked}) shows all three stress-test dimensions tracked against the retained signal information upon SF-degradation, whereas Part (\subref{fig:levels-subset}) highlights
  the transfer gap (see Section \ref{sec:dim3}) indicating a lack of SF-invariance in TSFMs.}
  \label{fig:disentangle}
  \vspace{-8pt}
\end{figure*}

We perform experiments on 3 publicly available datasets.
In HHAR \cite{stisen2015smart} we use accelerometer and gyroscope data from Samsung ($100$\,Hz) and LG ($200$\,Hz) watch to perform $6$-class human activity recognition.  
In CWRU bearing \cite{smith2015rolling} we use drive-end vibration data at $12$kHz and $48$kHz SFs to perform fault severity classification.  
In MIT-BIH \cite{moody2001impact} we use pre-segmented beats at $125$ Hz to perform $5$-class AAMI arrhythmia beat classification.
Since beats are pre-segmented, Dimension 2 forecasts each beat's final 32 samples from its prefix.
We anchor our discussion on Table~\ref{tab:stress} and Fig.~\ref{fig:disentangle} results.
\vspace{-6pt}
\subsection{Interesting properties of FSI, FDI, and CDI}
\label{subsec:metric-properties}
FSI is a non-negative and bounded metric in the range $[0,\sim1.8]$.
The cosine distance in FSI is measured within TSFM's own embedding space.
Hence, its magnitude is affected by model's embedding dimensionality, anisotropy, and multi-channel pooling strategy \cite{razzhigaev2024shape}, making FSI incomparable across TSFMs.
Still, FSI remains useful as it is computationally cheap and label-free.
FDI and CDI, on the other hand, are performance-driven metrics, and hence are comparable across TSFMs. 
Unlike FSI, FDI is unbounded and observed to be heavy-tailed (observed values run from $-0.011$ to $5.692$
), whereas CDI is bounded and carries practical significance for many downstream tasks.  
Since FSI is comparable only within a TSFM, we correlate FSI and CDI per TSFM over the seven settings in Table \ref{tab:stress}. 
They are strongly correlated for MOMENT, Moirai-MoE, and TimesFM with Pearson $r=0.99$, $0.97$, and $0.85$, respectively.
FDI, in contrast, is observed to be uncorrelated with both FSI and CDI.
Small negative FDI values (e.g., -0.011, TimesFM on Bearing 48 kHz) indicate that band-limiting slightly helps remove high-frequency content that is difficult to forecast. 
\vspace{-7pt}
\subsection{Do the indices track the information the signal loses?}
\label{subsec:info-loss}

Figure~\ref{fig:info-stacked} plots per-dimension discrepancy curves (cosine distance $\delta$ in row 1, excess nRMSE in row 2, and macro-$F_1$ drop in row 3) against the sampling frequency drop $\rho = 1 - K/F$, and compares them with the fraction of spectral information retained by the signal in terms of retained spectral energy ($R$) and spectral Jensen-Shannon similarity ($1-\text{JSD}$). 
Overall, we observe that these three discrepancy measures do not begin to degrade at the same point. 
Specifically, the embedding drift (row 1) responds first and grows smoothly while $R$ and $1-\text{JSD}$ are still close to 1.
On the other hand, forecasting and classification degradation (rows 2 and 3) stay negligible across a wide range of $\rho$ and then fall sharply past a threshold, which aligns with the point at which the retained-information curves start to collapse.
These patterns reveal that each dimension relates to information loss in fundamentally different ways. 
The embedding shift measures the model's sensitivity to a change in input, so it responds early and grows continuously. 
The downstream indices, however, behave differently: they remain stable across large SF drops and degrade only when the spectral content critical for their task is lost.
\vspace{-16pt}
\subsection{Are TSFMs SF-invariant? How do they compare?}
\label{subsec:tsfm-choice}
Fig. 2b compares $F_1^{\mathrm{tr}}(\rho)$ (native-SF probe) against $F_1^{\mathrm{orc}}(\rho)$ (probe at each $\rho$), and colors the transfer gap $\tau(\rho)$.
By definition, the transfer gap reflects a lack of SF-invariance rather than the loss of task-critical signal information.
For an ideal SF-invariant TSFM, this gap should be close to zero.
However, in all (TSFM $\times$ dataset) configurations, we observe a significant transfer gap, which suggests that there remains scope for TSFMs to be more SF-invariant.


Since FSI is not comparable across TSFMs (see Section \ref{subsec:metric-properties}) and FDI is undefined for MOMENT, we compare TSFMs using CDI and the transfer gap, the metrics which are the most practically relevant.  
Averaged over the seven dataset settings, we empirically observe MOMENT and TimesFM to be the most SF-invariant:
MOMENT has the lowest downstream degradation (CDI $=0.112$) and the smallest transfer gaps on the HHAR gyroscope streams, while TimesFM is a close second (CDI $=0.114$) and is the most stable on oversampled high-rate signals, with the lowest CDI on Bearing 48\,kHz ($0.037$) and a transfer gap that stays near zero until around $\rho \approx 0.9$. 
For the remaining two, Moirai-MoE ranked intermediate (with CDI $=0.127$), and Chronos-2 was observed to be the least SF-invariant, with the highest downstream degradation (CDI $=0.145$) and the widest transfer gap in most settings (see Figure~\ref{fig:levels-subset}). 
However, we also note that this SF-invariance is relative rather than absolute; no TSFM is truly invariant, and all four collapse on the near-critical Bearing 12\,kHz setting, where SF-degradation removes task-critical bandwidth and even the $F_1^{\mathrm{orc}}(\rho)$ drops. 
\vspace{-6pt}
\subsection{Can a contrastive alignment head close the transfer gap?}
\label{subsec:contrastive}
\begin{figure}[t]
  \centering
  \includegraphics[width=\linewidth]{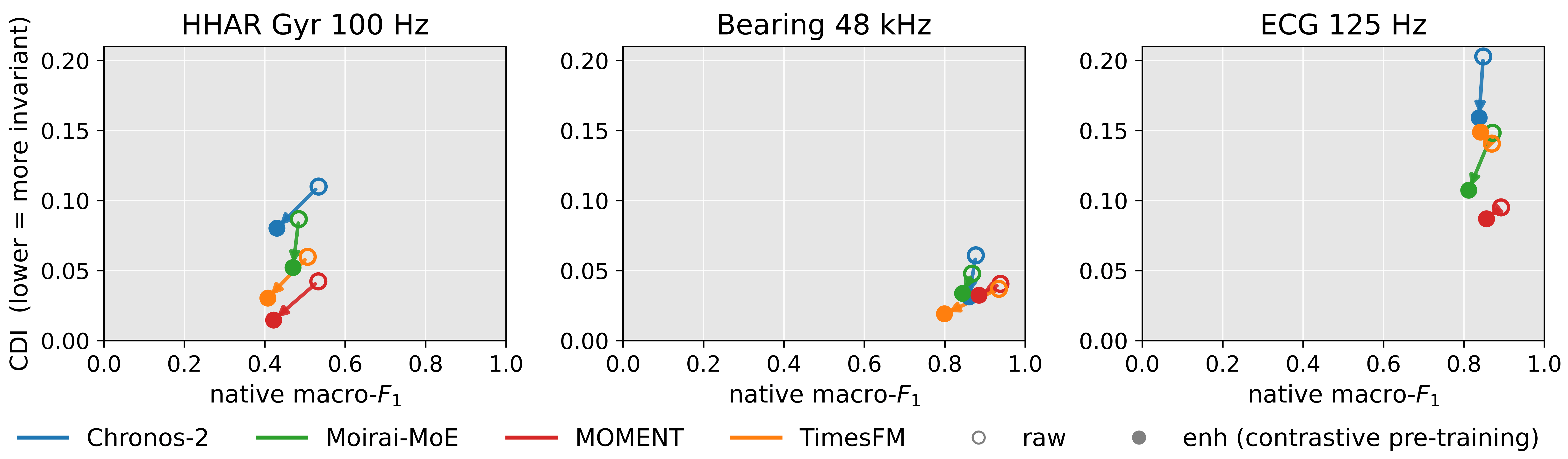}
  \caption{$F_1^{\mathrm{tr}}(0)$ vs.\ CDI with pre$\rightarrow$post contrastive alignment arrows.
  }
  \label{fig:perf-compact}
  \vspace{-14pt}
\end{figure}

The contrastive SF-alignment head (see Sec.~\ref{sec:adapter}) improved CDI in $26$ of the $28$ configurations (see Table~\ref{tab:stress}) by $40\%$ on average, with gains as large as $86\%$. 
While not necessarily optimal, this demonstrates that SF-invariance can indeed be improved.
However, there are two caveats to these improvements. 
First, the mean improvement is smallest for Moirai-MoE (26\%), one of the less SF-invariant models, and largest for TimesFM (46\%), one of the most invariant ones.
Second, the head causes some native-SF performance degradation across all TSFMs (Figure~\ref{fig:perf-compact}), revealing a trade-off between SF-invariance and native performance. 
The key challenge is thus to enhance SF-invariance without sacrificing native performance.
\vspace{-6pt}
\subsection{Insights on SF-invariance and practical guidelines}
\label{subsec:takeaways}
Taken together, our results suggest that SF-invariance is not an intrinsic property of TSFMs, and rather is jointly shaped by 3 factors. 
The first is how much gap exists between the native-SF and the signal's information bandwidth, since this gap influences what SF-reduction can remove. 
This is the reason why CWRU dataset's two tasks (at 12 and 48 kHz) behave differently (see Table \ref{tab:stress}), despite a majority of information loss happening $\sim6$ kHz in both.
The second is the factors of sensing modality and the downstream task, which together determine where in the spectrum the discriminative information lives. 
This is why, for the same activities, the gyroscope yields lower CDI than the accelerometer in 6 of 8 (TSFM$\times$SF) pairs, Chronos-2 being the exception (Table \ref{tab:stress}).
The third is the TSFM itself, which determines how commensurate its response is to SF-degradation and information bandwidth removed by it.
However, the impact of this factor is secondary.
A two-way variance decomposition of CDI values in
Table~\ref{tab:stress} attributes $94\%$ of their variance to dataset configuration and only $1.5\%$ to the TSFM, and still $65\%$ against $18\%$ when the Bearing $12$\,kHz setting is excluded.

These insights guide practitioners in using TSFMs for sensor applications.  
First, an estimate of how much SF-degradation the system can expect should be made
considering cheaper sensors, power saving, and heterogeneous devices.
Then, the native-SF should be set high enough to provide sufficient margin above the signal's information bandwidth to ensure that the expected SF-reduction stays within the safe range.
Second, the choice of TSFM should be driven mainly by task and engineering constraints (compute availability, model size, and throughput), because although the 4 TSFMs studied here differ in their SF-invariance, that difference matters less than the gap between native-SF and information bandwidth.

\vspace{-6pt}
\section{Conclusion \& Future Works}
\label{sec:conclusion}
\vspace{-4pt}
In this paper, we present, to our knowledge, the first study to place SF-invariance at the center of TSFM evaluation, quantifying it along embeddings (FSI), forecasting (FDI), and downstream classification (CDI) for 4 TSFMs. 
Our results show that TSFMs are not inherently SF-invariant; however, the invariance is governed more by the gap between native-SF and task-critical information bandwidth than the choice of TSFM itself.
These findings turn ``\textit{is this TSFM SF-invariant?}'' into a measurable, signal-aware question and guide practitioners in anticipating when a TSFM will transfer reliably across SFs. 
This study can be extended to more TSFMs and sensing modalities for more robust generalization. 
Moreover, our analysis acts as a motivation for training TSFMs to be more SF-invariant.

\section*{Acknowledgment}
The authors used Claude Code for implementation of experiments as well as Claude Cowork for grammar enhancement and editing of the paper.
The authors have verified the code as well as the revised write-up of the paper and take full responsibility of them. 

\section*{Compliance with Ethical Standards}
This research used publicly available datasets and did not involve any new studies with human participants or animals. 
No funding was received for conducting this study. The authors have no relevant financial or non-financial interests to disclose.
The authors declare that they have no conflict of interest.



\bibliographystyle{IEEEbib}
\bibliography{strings,refs}

\end{document}